\documentclass[runningheads]{llncs}
\usepackage[T1]{fontenc}
\usepackage{graphicx,verbatim}
\usepackage{todonotes}
\usepackage{amsmath}
\usepackage{amsfonts}
\usepackage{booktabs}
\usepackage{tabularx}
\usepackage[acronym,toc]{glossaries}

\newacronym{INRS}{INRS}{Implicit Neural Representation of Shapes}
\newacronym{INR}{INR}{Implicit Neural Representation}
\newacronym{UltrON}{UltrON}{Ultrasound Occupancy Network}
\newacronym{ON}{ON}{Occupancy Network}
\newacronym{SDF}{SDF}{Signed Distance Field}
\begin{document}
\title{UltrON: Ultrasound Occupancy Networks}
%
%
\author{Magdalena Wysocki\inst{1, 3}\orcidID{0009-0008-8239-1629} \and
Felix Duelmer\inst{1, 3}\orcidID{0009-0000-1702-1746} \and
Ananya Bal\inst{2}\orcidID{0000-0002-9592-1085} \and
Nassir Navab\inst{1, 3}\orcidID{0000-0002-6032-5611} \and
Mohammad Farid Azampour\inst{1, 3}\orcidID{2222--3333-4444-5555}}
%
\authorrunning{M. Wysocki et al.}
%
\institute{Chair for Computer Aided Medical Procedures (CAMP) \\ Technical University of Munich, Boltzmannstr. 3, 85748 Garching, Germany 
\email{\{magdalena.wysocki, felix.duelmer, nassir.navab, mf.azampour\}@tum.de} \and
Robotics Institute \\ Carnegie Mellon University, 5000 Forbes Avenue Pittsburgh, PA 15213, USA
\email{abal@andrew.cmu.edu} \and
Munich Center for Machine Learning (MCML), Munich, Germany
}
\maketitle              
\begin{abstract}
In free-hand ultrasound imaging, sonographers rely on expertise to mentally integrate partial 2D views into 3D anatomical shapes.
Shape reconstruction can assist clinicians in this process. 
Central to this task is the choice of shape representation, as it determines how accurately and efficiently the structure can be visualized, analyzed, and interpreted.
Implicit representations, such as SDF and occupancy function, offer a powerful alternative to traditional voxel- or mesh-based methods by modeling continuous, smooth surfaces with compact storage, avoiding explicit discretization.
Recent studies demonstrate that SDF can be effectively optimized using annotations derived from segmented B-mode ultrasound images. 
Yet, these approaches hinge on precise annotations, overlooking the rich acoustic information embedded in B-mode intensity. 
Moreover, implicit representation approaches struggle with the ultrasound's view-dependent nature and acoustic shadowing artifacts, which impair reconstruction.
To address the problems resulting from occlusions and annotation dependency, we propose an occupancy-based representation and 
introduce \gls{UltrON} that leverages acoustic features to improve geometric consistency in weakly-supervised optimization regime. 
We show that these features can be obtained from B-mode images without additional annotation cost.
Moreover, we propose a novel loss function that compensates for view-dependency in the B-mode images and facilitates occupancy optimization from multiview ultrasound.
By incorporating acoustic properties, \gls{UltrON} generalizes to shapes of the same anatomy.
We show that \gls{UltrON} mitigates the limitations of occlusions and sparse labeling and paves the way for more accurate 3D reconstruction. 
Code and dataset is available at https://github.com/magdalena-wysocki/ultron.


\end{abstract}
\section{Introduction}
\begin{figure}[h]
\includegraphics[width=\textwidth]{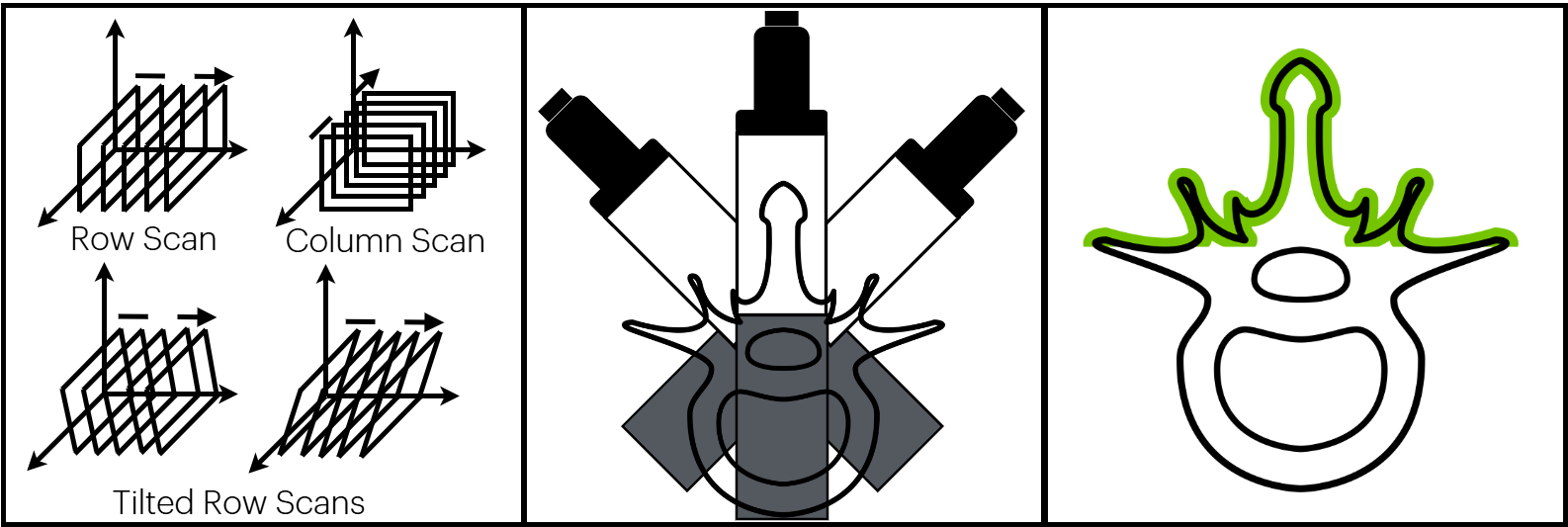}
\caption{(a) The method uses multiview ultrasound scans. We use row and column scans as proposed in RoCoSDF~\cite{chen2024rocosdf} and tilted scans as proposed in Ultra-NeRF~\cite{wysocki2024ultra}. (b) Occlusions in ultrasound B-mode imaging create partial observations. The regions in acoustic shadow (gray) are undefined. (c) Because of the occlusions we can define only the partial shape (green).}
\label{fig:input_data}
\end{figure}
In conventional free-hand ultrasound imaging, sonographers must rely on their expertise to mentally reconstruct the 3D shape of organs or structures from partial 2D views in order to extract the necessary anatomical information for diagnosis or intervention.
Shape reconstruction in medical ultrasound, can help clinicians in this task by enhanced visualization of a target structure.
In the field of computer vision, objects can be represented through a diverse array of shape representations.
Among them, implicit shape representations provide a method for representing 3D shapes as an implicit function defining the surface of an object. 
Unlike conventional representations such as meshes, point clouds, or voxel grids, implicit representations map spatial coordinates to a continuous function that describes the shape~\cite{Michalkiewicz_2019_ICCV}. 
\gls{INRS} is a deep-learning-based technique for approximating implicit shape functions. 
The key concept is that the function is parameterized by a neural network.
\gls{INRS} inherently provides a continuous representation, allowing it to capture fine details at any resolution without increasing memory usage. 
These representations define inherently smooth surfaces, making them useful for tasks requiring seamless shape transformations~\cite{yang2022implicitatlas}. 
Moreover, they enable learning families of shapes, facilitating the generation of novel shapes~\cite{yang2024generating},~\cite{kong2024sdf4chd}.
Among \gls{INRS} methods, Deep \gls{SDF}~\cite{park2019deepsdf} and \gls{ON}~\cite{mescheder2019occupancy} are two widely used approaches. 
In medical ultrasound, recent methods\textemdash UNSR~\cite{chen2024neural} and RoCoSDF~\cite{chen2024rocosdf} \textemdash leverage \gls{SDF}-based neural networks to represent anatomical structures from B-mode images. 
While UNSR relies on single-view ultrasound data, RoCoSDF demonstrates that incorporating multiple ultrasound views leads to more precise 3D shape reconstruction.

Multiview ultrasound scanning, as shown in Fig~\ref{fig:input_data}(a) combines perspectives from various angles, providing a more complete spatial representation and better visualization of complex anatomical structures. 
However, non-tomographic modalities like ultrasound B-mode imaging are inherently view-dependent and susceptible to acoustic shadowing occlusions, yielding only fragmented observations of the volume~\cite{gafencu2024shape}. 
The direction-dependent characteristics of ultrasound imaging create additional challenges for learning a shape representation, as scanning the same target from different directions can result in varying  signal intensities in the same spatial location.
Moreover, occlusions, visualized in Fig~\ref{fig:input_data}(b), make the regions under acoustic shadows undefined, therefore the shape can be only defined in the observed space (Fig~\ref{fig:input_data}(c)).
Since UNSR and RoCoSDF rely on densely annotated B-mode images for accuracy, they may encounter issues with occlusions, annotation errors, and partial annotations, which are common in B-mode imaging.

In this paper we propose \gls{UltrON} a novel \gls{INRS} for ultrasound that integrates acoustic features obtained from B-mode intensities into the representation to address the issues of annotation cost and accuracy.
Our contributions can be summarized as followed:
\begin{itemize}
    \item We demonstrate that by using the information in the B-mode intensities and without additional labels, the proposed method reduces the supervision required for learning INRS from ultrasound scans by 90\%.
    \item We tackle the problem of partial observations due to occlusions, by introducing an attenuation-compensated loss function. This enables optimization directly from multiview annotations.
    \item By integrating acoustic features into the occupancy function, the proposed method generalizes effectively, learning INRS across different volumes of the same anatomical structure.
\end{itemize}

\section{Method}
\begin{figure}[t]
\label{fig:method}
\includegraphics[width=\textwidth]{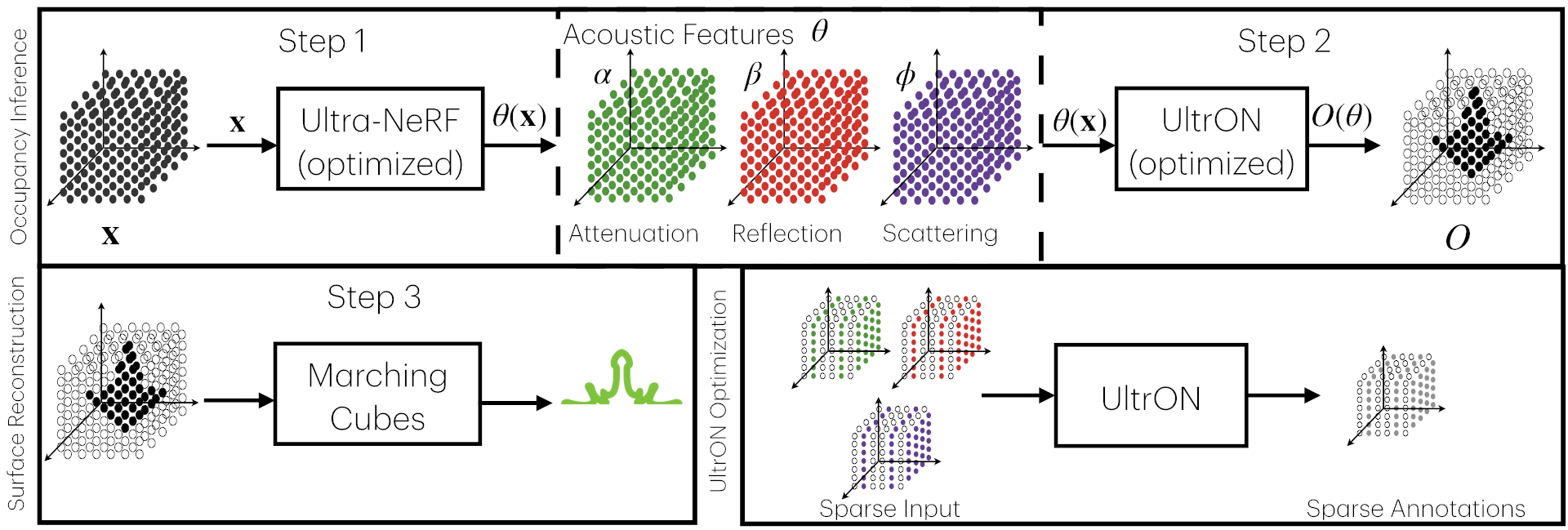}
\caption{Overview of the proposed three-step approach for medical shape representation and surface reconstruction from multiview ultrasound scans. First, a neural field of tissue-specific acoustic properties $\boldsymbol{\theta}$—attenuation ($\alpha$), reflection ($\beta$), and scattering ($\phi$)—is optimized. Second, an \gls{UltrON} is optimized to map these properties to occupancy ($o(\boldsymbol{\theta})$). Finally, the optimized occupancy network is sampled, and the Marching Cubes [3] algorithm is applied to extract the 3D mesh. Ultra-NeRF is optimized using the full 3D volume as supervision, whereas UltrON is optimized using sparse 2D ultrasound annotations converted to occupancy.}\label{fig1}
\end{figure}
\subsection{Overview}
Our objective is to reconstruct the surface of a medical shape from multiview ultrasound scans. 
As shown in Fig.~\ref{fig:method}, the proposed method follows a three-step approach. First, we optimize a neural field of tissue-specific acoustic properties (attenuation, reflection, scattering). Second, we optimize an occupancy network that maps these properties to occupancy. Finally, we sample the optimized occupancy network and apply Marching Cubes~\cite{lorensen1998marching} algorithm to extract the 3D mesh. 
\subsection{Neural Field of Acoustic Features}
Our approach is based on the insight that the same tissue type exhibits consistent acoustic features, therefore if one knows distribution of these features for a specific tissue it can be used to localize this tissue in space. 
In B-mode images, we observe only pixel intensities as effects of acoustic features, not the features themselves.
To learn these features from multiview B-mode scans we employ Ultra-NeRF.
Ultra-NeRF is a neural rendering framework that allows synthesis of B-mode images from unobserved viewing points. 
Additionally it provides a distribution of the three acoustic features—namely, attenuation ($\alpha$), reflection ($\beta$), and scattering ($\phi$)-in space. 
However, the mapping between the distribution and the tissue type is unknown.
To find this mapping we propose using an occupancy-based method.
Since these features are approximately homogeneous within the same tissue type, we demonstrate that optimizing this mapping requires fewer annotations compared to direct coordinate-to-occupancy mapping.
\subsection{Ultrasound Occupancy Network}
Occupancy function $o$ of a 3D object is defined as a function that for every 3D point $\boldsymbol{x} \in \mathbb{R}^3$ maps this point to an occupancy value: 
\begin{equation}
o : \mathbb{R}^3 \to \{0,1\}
\end{equation}
For a single 3D object, \gls{ON}~\cite{mescheder2019occupancy} is a neural implicit representation used to model 3D object by predicting the occupancy probability of points in continuous space. 
Given a spatial coordinate \( \boldsymbol{x} \in \mathbb{R}^3 \), an occupancy network approximates the occupancy function:
\begin{equation}
f_{\omega} : \mathbb{R}^3 \to [0,1]
\end{equation}
where \( f_{\omega}({\boldsymbol{x}}) \) represents the probability that the point $\boldsymbol{x}$ is inside the object.
In this paper, we introduce \gls{UltrON}, which extends the standard \gls{ON} by integrating acoustic information into the representation. 
Specifically, instead of relying solely on spatial coordinates, the ultrasound occupancy function $o_u$ is reformulated as
\begin{equation}
o_u : \mathbb{R}^d \to \{0,1\},
\end{equation}
\begin{equation}
o_u (\boldsymbol{\theta}(\boldsymbol{x})) =  
\begin{cases}
    1 & \text{if occupied} \\
    0 & \text{otherwise.}
\end{cases}
\end{equation}
We approximate $o_u$ by the network $f_{\omega}$ and therefore $f_{\omega} (\boldsymbol{\theta}(\boldsymbol{x})))$ represents the probability that the point \( \boldsymbol{x} \) is inside the object given acoustic properties at the point $\boldsymbol{x}$ and $\boldsymbol{\theta}(\boldsymbol{x}) \in \mathbb{R}^d$ is the vector of acoustic properties at point $\boldsymbol{x}$ .
Since acoustic properties are largely homogeneous within a given tissue type but vary across different tissues, incorporating this information provides a more structured representation of anatomical regions. 
\subsection{Attenuation-compensated Optimization}
Ultrasound is view-dependent, therefore during the training we account for acoustic attenuation when comparing the network output with ground truth labels, ensuring accurate tissue identification across different imaging angles.
To this end, we define the loss function based on the binary cross-entropy (BCE) loss that considers effects of attenuation along the propagation path of the ultrasound beam. 
The rationale behind this is that some regions within the volume of interest may or may not be observed depending on the probe's position and orientation since occlusions can prevent the signal from reaching certain points, leading to incomplete or inaccurate observations. 
The resulting loss at a given point \( \mathbf{x} \) is computed as follows:
\begin{equation}
\mathcal{L}(\boldsymbol{x}) =  - \left[ y(\boldsymbol{x}) \cdot \log( T(\boldsymbol{x})\cdot f_{\omega} (\boldsymbol{\theta}(\boldsymbol{x}))) + (1 - y(\mathbf{x})) \cdot \log(1 - T(\boldsymbol{x}) \cdot f_{\omega} (\boldsymbol{\theta}(\boldsymbol{x})) \right]
\end{equation}
where $f_{\omega} (\boldsymbol{\theta}(\boldsymbol{x}))$ is the occupancy function,
$y(\boldsymbol{x})$ is the ground truth label, and
$T(\boldsymbol{x})$ is the transmittance function at point $\boldsymbol{x}$. 
$T(\boldsymbol{x})$ accounts for the transmission of the acoustic signal based on the probe position and orientation. 
It reflects occlusions and the visibility of the volume from the given probe viewpoint and is defined as:
\begin{equation} \label{eg:remaining_intensity_short}
T(\boldsymbol{x}) = T(0) \cdot \exp^{-{\int_{0}^{x-\epsilon}{\beta(n)dn}}}\cdot \exp^{-{\int_{0}^{x-\epsilon}{\alpha(n)dn}}}
\end{equation}
To compute $T(\boldsymbol{x})$ we use the ray casting and integrate attenuation $\alpha$ and reflection $\beta$ along propagation path of the ultrasound beam from transducer to the current position $\boldsymbol{x}$ within the volume. 
Similar integration of reflection and attenuation has been proposed in ~\cite{yesilkaynak2024ultrasound},~\cite{wysocki2024ultra},~\cite{salehi2015patient}.
\section{Experiments and Results}
\begin{figure}[t]
\includegraphics[width=\textwidth]{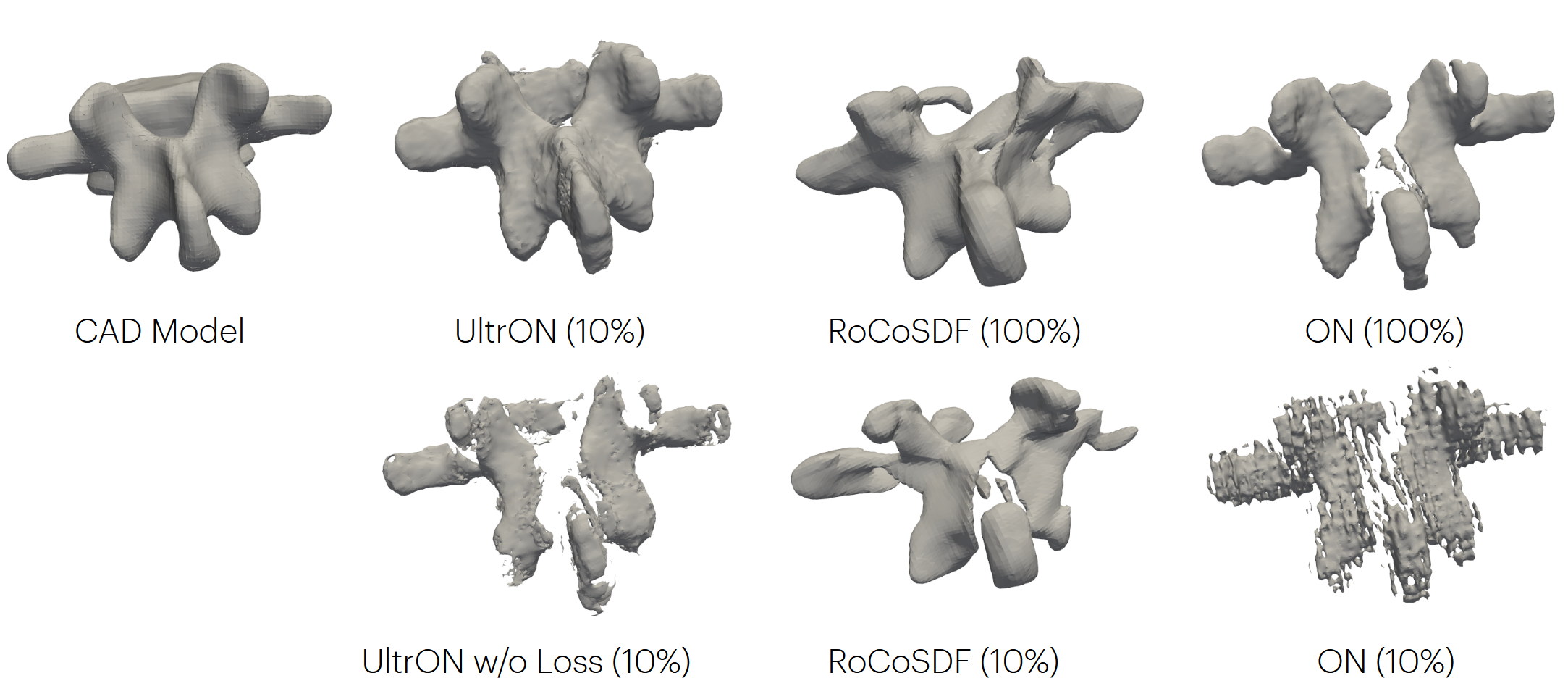}
\caption{Visualization of an example if reconstruction of L3 vertebra model from different INRS optimized with dense (100\%) and weak supervision (10\%). We observe that UltrON optimized on 10\% of manual annotations preserves topology more accurately than RoCoSDF and \gls{ON} optimized on 100\% annotations. This shows that UltrON compensates errors in annotations naturally occurring due to tracking errors, occlusions and human errors. Comparison between UltrON with (upper) and UltrON without (bottom) the attenuation-compensated loss shows that the loss compensates acoustic shadows in multiview ultrasound of a highly reflective structure.} 
\label{fig:results_all}
\end{figure}
\subsection{Data}
Four CAD vertebra models of lumbar spine vertebrae (L2, L3x2, L4), created from the VerSe dataset~\cite{sekuboyina2021verse}, are used as phantoms for 3D printing. 
These models are then used to create ballistic gelatin-based phantoms with paper pulp that mimic soft tissue~\cite{jiang2021autonomous}, offering better realism compared to models in a water bath.
Similar to RoCoSDF, we performed multiview scanning that includes row and column scans, and in addition tilted scans as well. 
multiview scanning integrates multiple viewpoints as illustrated in Fig~\ref{fig:input_data}(a).
The tilted scans involve adjusting the probe by -10 and +10 degrees from the standard row scan position.
For acquisition, we use a probe mounted on a robotic arm. 
The robotic tracking system is calibrated to account for the offset caused by the probe's attachment to the arm.
After data acquisition, the B-mode images are manually segmented. 
This segmentation is then converted to occupancy data, where 0 is background label and 1 is bone label.
Using the poses corresponding to the B-mode images, we compute the voxel positions in space and normalize them to fit within a unit cube.
For training RoCoSDF, we follow the procedure outlined in \cite{chen2024rocosdf} to generate the point clouds. 
For Ultra-NeRF, we utilize the poses and B-mode images as described in \cite{wysocki2024ultra}.
\subsection{Implementation Details}
$f_{\omega}$, the occupancy function in \Gls{UltrON}, and Ultra-NeRF, use the same architecture which consists of 8 fully-connected layers with 128 hidden channels and with the skip connection at the fourth layer. 
We use ReLU activation layers except for the last layer.
We follow 2-stage optimization.
In the first stage, we optimize Ultra-NeRF for 75k iterations, then we optimize \gls{UltrON} for 50k iterations.
We use the Adam optimizer with 0.0001 learning rate and exponential decay. 
For training Ultra-NeRF we use regularized version of the method~\cite{yesilkaynak2024ultrasound} and default settings. 
We apply positional encoding method presented in NeRF~\cite{mildenhall2021nerf} to the input of the network. 
Our network is implemented using Pytorch and trained on NVIDIA RTX 3090 GPU with 24 GB memory. 
Smoothing is applied to the occupancy, and after smoothing the threshold for Marching Cubes is set to 0 to extract the zero-level-set of the surface.
For the smoothing and Marching Cubes we use PyMCubes\footnote{https://github.com/pmneila/PyMCubes} implementation. 
For \gls{ON}, we use the same architectur as $f_{\omega}$ but we change the network input to coordinates.
\subsection{Evaluation Method}
We compare the method to RoCoSDF, the state-of-the-art method in INRS for ultrasound imaging, and vanilla \gls{ON}.
Four evaluation metrics are used to assess reconstruction quality: Chamfer Distance (CD), Hausdorff Distance (HD), Mean Absolute Distance (MAD), and Root Mean Square Error (RMSE). 
These metrics are computed by calculating the distances between points randomly sampled from the reconstructed mesh and the corresponding CAD models. 
To further asses the reconstructed volumes we compare the reconstructed surfaces visually with respect to CAD models.
\begin{table}[t]
    \caption{Comparison of the shape reconstruction based on different representations optimized with dense (100\%) and weak supervision (10\%) as measured by Chamfer Distance (CD), Hausdorff Distance (HD), Mean Absolute Deviation (MAD), and Root Mean Square Error (RMSE).
    We observe that UltrON with 10\% annotations outperforms RoCoSDF trained on the whole available data by a margin of 26\% as measured by CD.}
    \centering
    \begin{tabularx}{\linewidth}{lXXXX}
        \toprule
        Method (supervision) & CD (mm)↓ & HD (mm)↓ & MAD (mm)↓ & RMSE (mm)↓\\
        \midrule
        RoCoSDF~\cite{chen2024rocosdf}(100\%) & 2.98  $\pm$ 0.03 & 9.79  $\pm$ 0.02 & 2.53  $\pm$ 0.02 & 3.67  $\pm$ 0.03 \\
        \midrule
        ON (100\%) & 2.91  $\pm$ 0.03 & 9.23  $\pm$ 0.02 & 2.60  $\pm$ 0.03 & 3.70  $\pm$ 0.03\\

        \midrule
        UltrON w/o Loss (10\%) & 3.25  $\pm$ 0.12 & 9.30  $\pm$ 0.06 & 2.82  $\pm$ 0.12 & 3.98  $\pm$ 0.15 \\
        UltrON (10\%) & \textbf{2.22  $\pm$ 0.02} & \textbf{7.98  $\pm$ 0.04} & \textbf{1.67  $\pm$ 0.02} & \textbf{2.69  $\pm$ 0.03}\\
        UltrON (5\%) & 2.36  $\pm$ 0.03 & 8.04  $\pm$ 0.05 & 1.85  $\pm$ 0.03 & 2.88  $\pm$ 0.03 \\
        \bottomrule
    \end{tabularx}
    \label{tab:results}
\end{table}
\subsection{Qualitative and Quantitative Results}
\begin{figure}[t]
\includegraphics[width=\textwidth]{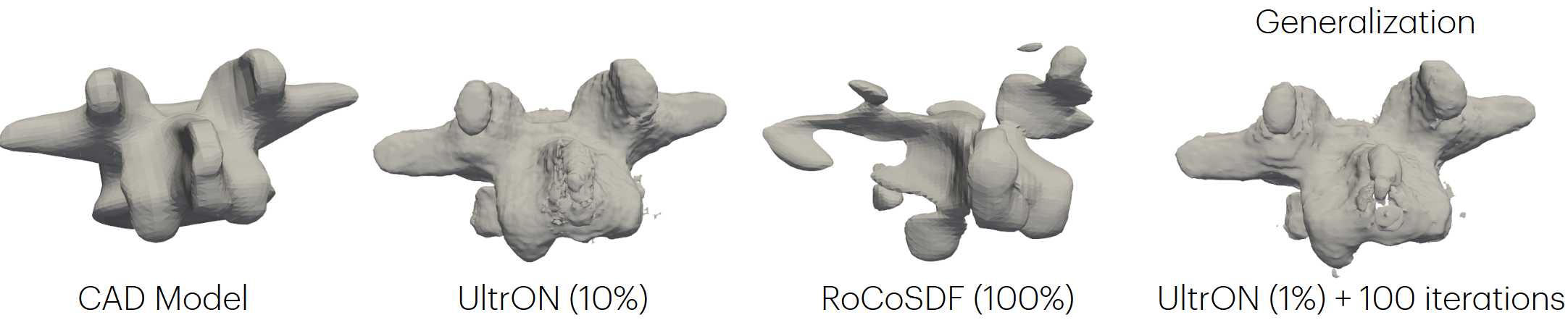}
\caption{Visualization of the reconstructed vertebra shapes with UltrON and RoCoSDF in the presence of larger annotation errors and visualization of generalization to the new shape (trained on L3 and fine-tuned on L2).} 
\label{fig:robustness_to_errors_result}
\end{figure}

As shown in Table~\ref{tab:results}, \gls{UltrON} improves surface reconstruction by 26\% (CD) over RoCoSDF and coordinate-based \gls{ON} and requires 90\% less annotations to achieve this performance.
Quantitative results in Fig.~\ref{fig:results_all} show that \gls{UltrON} better preserves topology even in sparse data regimes since it is using dense information about distribution of the acoustic features within the volume.
We observe that \gls{UltrON} is moreover more robust to errors in input data resulting from annotation errors and in the misalignment of the input data.
We observe that with these errors RoCoSDF fails to preserve surface topology whereas using acoustic features directly helps \gls{UltrON} to correct these errors.
We show an example qualitatively in Fig~\ref{fig:robustness_to_errors_result}.
\subsubsection{Loss Ablation}
To visualize the importance of the attenuation compensation we perform ablation on the loss. 
In Table~\ref{tab:results} and Fig~\ref{fig:results_all} we show that using a standard BCE loss decreases performance of the method since the optimization does not compensate the occlusions resulting from attenuation.
This, as we can see in the Fig~\ref{fig:results_all} results in errors in the topology due to occlusions.
\begin{table}[]
    \caption{Performance on generalization to a new shape of the same anatomy based on Chamfer Distance (CD), Hausdorff Distance (HD), Mean Absolute Deviation (MAD), and Root Mean Square Error (RMSE).}
    \centering
    \begin{tabularx}{\linewidth}{lXXXX}
        \toprule
        Supervison \% + \# iter. \ \  & CD (mm)↓ & HD (mm)↓ & MAD (mm)↓ & RMSE (mm)↓ \\
        \midrule
        w/o fine-tunning  & 3.44  $\pm$ 0.02 & 13.23  $\pm$ 0.07 &  1.79  $\pm$ 0.02 & 2.67  $\pm$ 0.03 \\
        1\% + 100 iter  & 2.44  $\pm$ 0.02 & 7.91  $\pm$ 0.04 &  2.00  $\pm$ 0.02 & 2.94  $\pm$ 0.03 \\
        \bottomrule
    \end{tabularx}
    \label{tab:generalization}
\end{table}
\subsubsection{Shape Generalization}
To test the generalization, we need to account for real-world variations, as well as variations resulting from the ill-posed nature of optimizing a neural field. To this end, fine-tuning the network to a new shape of the same anatomical structure is necessary.
In this process, the last two layers of the network are frozen, and only 1\% of the labels are used.
The fine-tuning procedure takes approximately 5 seconds to complete (100 iterations).
In Table~\ref{tab:generalization} we show that the reconstruction quality as measured by the four reconstruction metrics is comparable to the full training.
Fig~\ref{fig:robustness_to_errors_result} presents this observation quantitatively.
\section{Conclusion}
We present \gls{UltrON}, a novel approach that integrates acoustic features from B-mode intensities into a representation of occupancy.
We show that with $90\%$ fewer annotations, \gls{UltrON} provides a representation that can enhance reconstruction accuracy by $26\%$. 
Moreover, we introduce an attenuation compensated loss function that facilitates optimization directly from multiview annotations and  tackles the problem of partial observations due to occlusions in multiview ultrasound.
Finally, we demonstrate that defining occupancy function as a mapping between acoustic features and occupancy enables generalization to the same anatomy across different volumes with 1\% of the supervision and only requires 100 iterations of fine-tuning.
To further enhance the generalization of UltrON, one could consider incorporating shape priors into the representation~\cite{amiranashvili2024learning},~\cite{bastian2023s3m}.
We also believe that explicitly defining unmeasured regions and incorporating uncertainty quantification can enhance reconstruction quality, as demonstrated through visibility analysis~\cite{wysocki2023scan2lod3}. With these improvements, UltrON has the potential to facilitate the creation of realistic, patient-specific 3D anatomical models that could be utilized by both clinical practitioners and automated systems such as robotic ultrasound~\cite{jiang2023robotic}.
\begin{credits}
\subsubsection{\ackname} This work was supported by the HINAV project funded by the Bavarian State Ministry for Economic Affairs, Regional Development and Energy. 

\end{credits}
\bibliographystyle{splncs04}
\bibliography{Paper-2979}
\end{document}